\documentclass[runningheads]{llncs}
\pdfoutput=1
\usepackage{graphicx}
\usepackage{booktabs}
\usepackage{multirow}
\usepackage{siunitx}

\begin{document}
\title{Dynamic Adaptive Spatio-temporal Graph Convolution for fMRI Modelling}

\author{Ahmed El-Gazzar\inst{1} \and
Rajat Mani Thomas\inst{1} \and
Guido van Wingen\inst{1}}
\authorrunning{A. El-Gazzar et al.}
%
\institute{{Amsterdam UMC, University of Amsterdam, \\
Department of Psychiatry, Amsterdam, The Netherlands \\}} 
\maketitle              
\begin{abstract}
The characterisation of the brain as a functional network in which the connections between brain regions are represented by correlation values across time series has been very popular in the last years. Although this representation has advanced our understanding of brain function, it represents a simplified model of brain connectivity that has a complex dynamic spatio-temporal nature. Oversimplification of the data may hinder the merits of applying advanced non-linear feature extraction algorithms. To this end, we propose a dynamic adaptive spatio-temporal graph convolution (DAST-GCN) model to overcome the shortcomings of pre-defined static correlation-based graph structures. The proposed approach allows end-to-end inference of dynamic connections between brain regions via layer-wise graph structure learning module while mapping brain connectivity to a phenotype in a supervised learning framework. This leverages the computational power of the model, data and targets to represent brain connectivity, and could enable the identification of potential biomarkers for the supervised target in question. We evaluate our pipeline on the UKBiobank dataset for age and gender classification tasks from resting-state functional scans and show that it outperforms currently adapted linear and non-linear methods in neuroimaging. Further, we assess the generalizability of the inferred graph structure by transferring the pre-trained graph to an independent dataset for the same task. Our results demonstrate the task-robustness of the graph against different scanning parameters and demographics.

\end{abstract}
\section{Introduction}
A major goal of neuroimaging studies is to develop predictive models to analyze the relationship between whole brain functional connectivity patterns and behavioural traits in order to better characterize brain function and dysfunction. Functional magnetic resonance imaging (fMRI) offers a promising window to investigate the brain activity by measuring the blood oxygenation level which offers a promising proxy to measure the neural activity of the brain. However, extracting the spatiotemporal features from fMRI scans is challenging  due to the high dimensionality of the data ($\sim$ 1 M voxels), low signal-to-noise ratio and limited availability of labeled datasets.   
A standard approach to overcome these challenges is modelling functional connectivity as the pairwise correlations of the time-series of pre-defined brain regions and then feeding these to a classifier either as input features directly\cite{vieira2017using} or as the underlying structure of a graph\cite{rubinov2010complex}. 
The study of the brain as a graph offers a natural representation of the underlying mechanisms of brain activity\cite{sporns2005human}. Together with the application of graph theory approaches, the field of network neuroscience has significantly deepened our understanding of brain function and dysfunction. Recently, with the progress of geometric deep learning, graph convolution networks (GCNs) are being exploited in the analysis of fMRI scans\cite{wanggraph,zhang2021functional}. A more befitting model for the dynamics of the brain are spatio-temporal GCNs (ST-GCNs)\cite{yan2018spatial}. \cite{gadgil2020spatio,azevedo2020towards} recently evaluated the application of ST-GCNs for fMRI analysis for age and gender classification. While these adaptions show a promising direction for modelling the high dimensional signal of the fMRI scan using a spatio-temporal model, the result metrics do not display a significant improvement from shallow non-linear or linear models applied on the flattened correlation matrix. These findings align with the rise of multiple research papers that debate the merits of applying advance deep learning techniques in neuroimaging for phenotype prediction even with scaling of the number of training samples\cite{schulz2020different,he2020deep}. We hypothesize that these results are attributed to (1) the use of engineered features such as time-series correlations to represent the data of functional parcellations of brain regions, which limits the modelling capability of the deep learning techniques even if a suitable network architecture is implemented. (2) A static graph structure assumes static brain connectivity which contradicts the dynamic nature of brain connectivity \cite{allen2014tracking} and its role in mapping to behavioural traits. (3) The adaption of exact neural network architectures that have shown previous success in other domains however are not well tailored to the nature of fMRI datasets such as the low temporal resolution and the scarcity of labelled samples.  
To address these issues, we propose a novel spatio-temporal graph convolution architecture with an adaptive graph structure learning. Our approach captures the spatio-temporal dependencies by combining graph convolutions and dilated 1D convolutions. Simultaneously, the graph structure learning module enables the inference of underlying dynamic graph structure. We evaluate our framework on the UKBiobank dataset for the tasks of sex and age classification form resting-state fMRI scans. We compare our results against different baselines as well as ablated versions of the model and we assess the robustness of the trained graph structure for the same task across an independent dataset. The remainder of this paper is ordered as follow: Section 2 describes the key components and the framework of the proposed architecture. Section 3 demonstrates the experiments conducted for the validation of the framework. Section 4 addresses generalizability of the inferred graph structures. Section 5 highlights the limitations of our work. Finally, Section 6 concludes this work and discusses the takeaways and future directions.   
\section{Methodology}

\subsection{Preliminaries}
We model the brain as $M$ directed dense graphs $\{G^{k} = (V,A^{k})\}_{k=1}^{M}$, $V$ are the nodes of the graph, where each node represent an anatomical region of the brain pre-defined using a template parcellation; $|V|=N$ is the number of regions in the template. $A^k$ $\in$ $R^{N \times N}$ represent a directed dense adjacency matrix initialized randomly. The signal on the graph is defined as $X$$\in$$R^{N \times T \times C}$ where $T$ is the number of sampled data-points from the scan and $C$ is the number of channels.

\subsection{Temporal lag correction}
The low temporal resolution and the existence of the temporal lag patterns in the fMRI signals was highlighted in previous studies\cite{mitra2014lag}. This behaviour can limit the model ability to capture the underlying spatio-temporal signal. To correct for this latency, we compute the first and second derivative of the global time-course signal of each brain region, stack them as 3 channels and apply a $1\times1$ convolution with a linear activation function before passing the signal to the spatio-temporal blocks. 

\subsection{Temporal Feature Extraction}
We adopt dilated 1D-convolution \cite{oord2016wavenet} as our temporal convolution layer (TCN) to capture a node’s temporal trends. Unlike \cite{oord2016wavenet} however, we opt for non-causal convolutions, because (i) our objective is learning global temporal representations for a classification task, rather than a generative or horizon prediction task, and (ii) fMRI signals are known to exhibit a temporal lag structure so access to future timepoints could be essential to effectively model intrinsic brain networks. We use gated activation similar to\cite{oord2016conditional}
\begin{equation}
    z = \text{tanh}(W_{k,f}*x)\odot \sigma(W_{k,g}*x)
\end{equation}
where $*$ denotes a convolution operator,$\odot$ denotes an element-wise multiplication operator, $\sigma$ is
a sigmoid function, $k$ is the layer index, $f$ and $g$ denote filter and gate, respectively, and $W_f$ and $W_g$ are the learnable filter and gate 1D convolutions. We empirically evaluate gated activation against rectified linear unites. We find that gated activation works better for modelling fMRI signal in terms of final metrics and loss convergence.

\begin{figure}[t!]
\includegraphics[width=\textwidth]{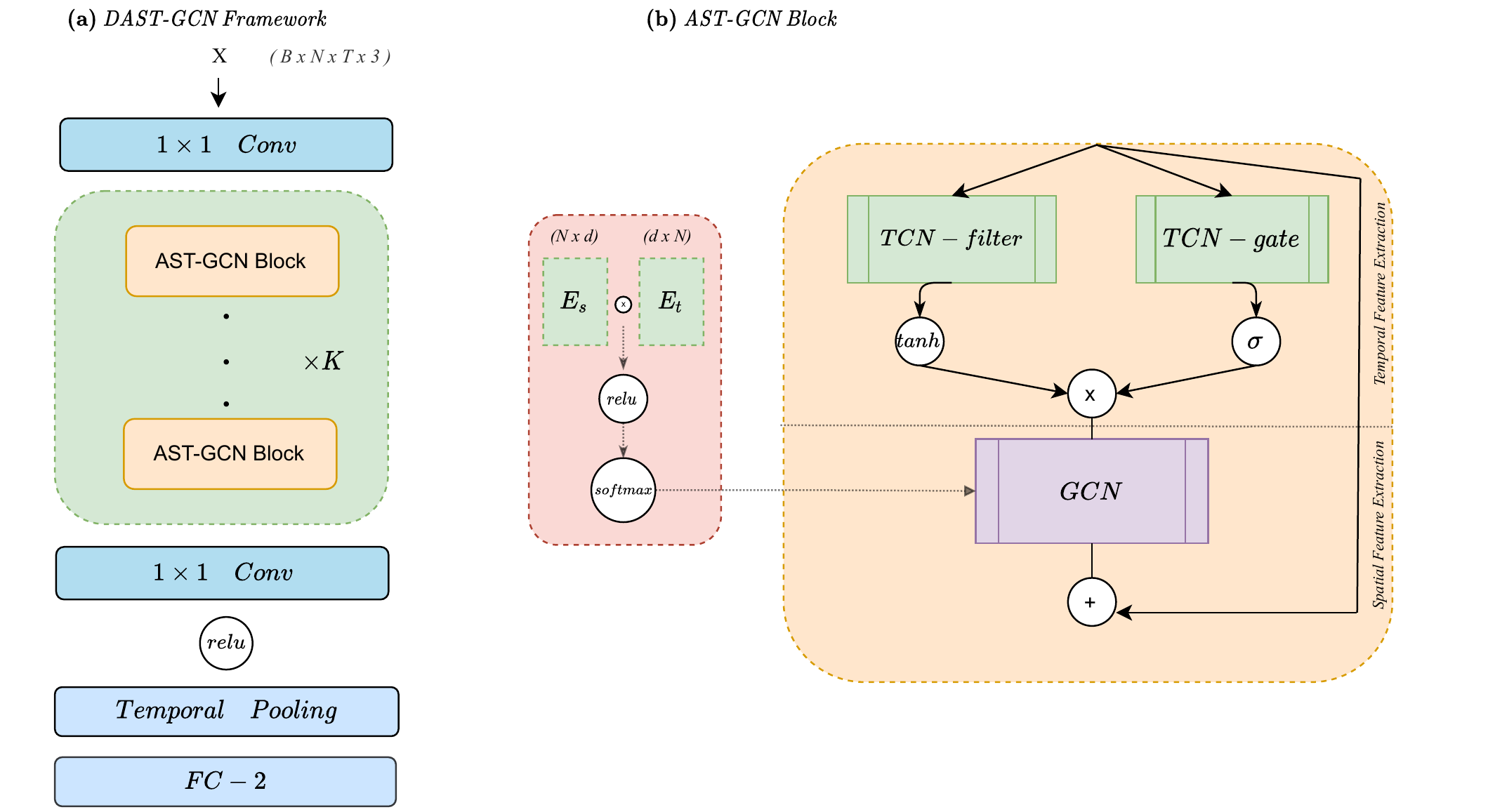}
\caption{Overview of the DAST-GCN architecture and its main components.} \label{fig1}
\end{figure}

\subsection{Spatial Feature Extraction}
To extract the spatial correlations of data with unstructured topologies, extending neural networks to process graph-structured data has attracted widespread attention\cite{wu2020comprehensive}. A series of studies has extended traditional convolution to model arbitrary graphs on spectral \cite{bruna2013spectral,defferrard2016convolutional} or spatial \cite{kipf2016semi,gilmer2017neural} domains. Spectral-based methods use a graph spectral filter to smooth the input signals of nodes. Spatial-based methods extract high-level representations of nodes by gathering feature information of neighbors. \cite{kipf2016semi} shows that graph convolution operation can be well-approximated by 1-st order Chebyshev polynomial expansion and generalized to high-dimensional GCN as:
\begin{equation}
    H = A^{'}XW
\end{equation}
where $A^{'}$ is the normalized adjacency matrix with self-loops. In neuroimaging, the most common approach to pre-define $A$ is a correlation metric between the timecourses. In this work, we inspire from previous studies on adaptive adjacency matrix learning \cite{wu2019graph,bai2020adaptive} and propose layer-wise graph structure learning module to capture the dynamic spatial dependency across different receptive fields. The normalized adjacency matrix can then be defined as:
\begin{equation}
    A^{'}_{k} = I_N + \text{Softmax}(\text{ReLU}(E_{s,k}E_{t,k}))
\end{equation}
$E_s$$\in$$R^{N \times d}$ and $E_t$$\in$$R^{d \times N}$ are trainable source and target dictionaries, where $d$ is a model hyper-parameter.
By multiplying $E_s$ and $E_t$, we derive the spatial dependency weights between the source nodes and the target nodes. We use the $ReLU$ and $Softmax$ activation functions to eliminate weak connections and normalize the adjacency matrix respectively. Finally, the identity matrix $I_N$ is added for the self loops.
\subsection{Framework of the model}

We present the framework of DAST-GCN in Figure 1. First a 1x1 convolution layer with shared weights is applied on each node signal independently to scale up the number of features. Followed by a stack of spatio-temporal blocks. A spatial-temporal block is constructed by  a gated temporal convolution layer (Gated TCN) with shared weights across the nodes, an Adaptive graph convolution layer (GCN) and a residual connection. By stacking multiple spatial-temporal layers, DAST-GCN is able to handle spatial dependencies at different temporal levels. Subsequently, Another 1x1 convolution is applied to reduce the node feature dimensions to a single temporal vector, followed by a temporal pooling layer and a dropout layer for regulization. Finally a fully connected layer with softmax activation is applied for the binary classification task.
\section{Experiments}

\subsection{Dataset}
We evaluated our pipeline on rs-fMRI scans of 6709 (3255 males / 3454 females, Age = 67.3 $\pm$ 7.2) participants of UKBiobank dataset. First, the provided pre-processed scans were registered to an MNI template followed by applying the AAL atlas \cite{tzourio2002automated} to parcellate the brain into N = 116 anatomical regions of interest (ROIs). The time-courses (TR=0.7s, T=490 timepoints) were averaged across the voxels of each ROI to obtain the signal for each node. We conducted binary sex classification and binary age classification for people above and under the age of 70 years to benchmark our framework.
\subsection{Experimental Setup}
We optimized the model for 200 epochs using Adam optimizer and binary cross-entropy loss with an initial learning rate of 0.001 and cosine scheduler with warm-up. A 5-fold cross-validation scheme was adapted for all the experiments. The models were trained on a Nvidia Tesla P100 GPU with a batch size of 32 and an average training time of 44 s/epoch.  After hyper-parameter search, we selected the number of spatio-temporal blocks $K$ = 3,  the embedding dimension of the node dictionaries $d$ = 10, the number of filters $f$ = 10 and the kernel size $ks$ = 3 for the temporal convolution layers across the model. The same configuration was selected for both tasks. Parameter sharing and the utilization of 1x1 convolutions across the model enables the development of an efficient model with 11205 trainable parameters. Our implementation is available open-source.\footnote{https://github.com/AhmedElGazzar0/DAST-GCN}

\subsection{Experimental Results}
\begin{figure}[t!]
\centering
\includegraphics[width=\textwidth]{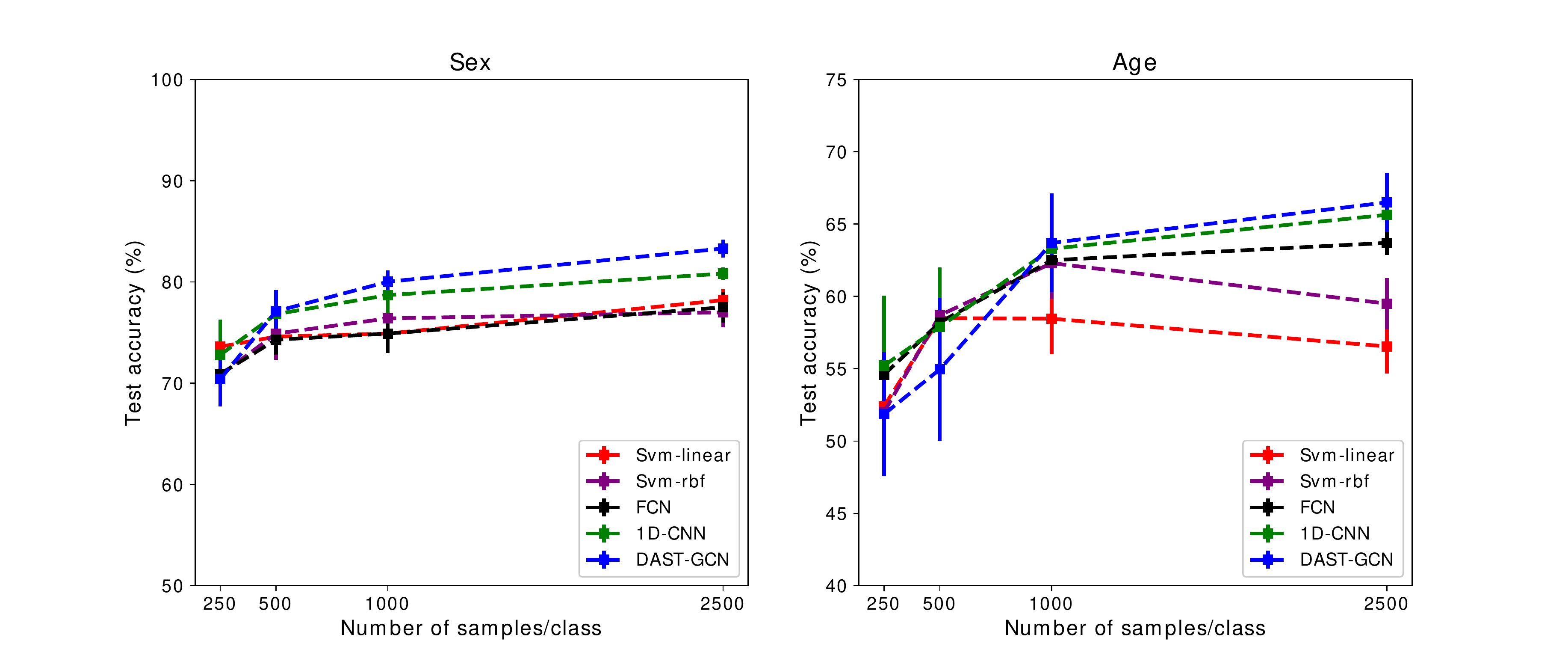}
\caption{5-fold scaling performance for DAST-GCN at sex and age classifications against baseline models under different numbers of samples per class. Also shown are the standard-deviation across folds as error-bars.} \label{fig2}
\end{figure}

We compared the performance of DAST-GCN against the following baseline models:
\begin{itemize}
    \item \textbf{SVM-linear}: a support vector machine with a linear kernel on the flattened Pearson's correlation matrix.
    \item \textbf{SVM-rbf} a support vector machine with a radial basis function kernel on the flattened Pearson's correlation matrix.
    \item \textbf{FCN:} a stack of fully connected layers on the flattened Pearson's correlation matrix.
    \item \textbf{BrainNetCNN\cite{kawahara2017brainnetcnn}:} a 2D CNN on Pearson correlation matrix.
    \item \textbf{1D-CNN\cite{el2019simple}:} a 1D CNN on the ROIs timecourses.
    \item \textbf{ST-GCN\cite{gadgil2020spatio}:} a static ST-GCN where the nodes features represent ROI timecourses and the edges are defined using thresholded Pearson's correlation.

\end{itemize}
For a fair comparison, we conducted a hyper-parameter search for the all the baselines models with the exception of \textbf{BrainNetCNN}\cite{kawahara2017brainnetcnn} and \textbf{ST-GCN}\cite{gadgil2020spatio} where we used the default parameters recommended by the authors. Moreover, to evaluate the efficiency of our architecture design, we created ablated versions of the model; \textit{DAST-GCN\_tlc:} without temporal lag correction module, \textit{DAST-GCN\_M=1:} only one adjacency matrix is trained throughout the model, \textit{DAST-GCN\_corr:} adjacency matrix is pre-computed as the mean Pearson's correlation matrix of the training samples, \textit{DAST-GCN\_undir:} a symmetric undirected adjacency matrix, i.e. $E_t$ $=$ $E^{T}_s$. A comparison of the results of the model against baselines and ablated versions are shown in Table 1.\\ The results show superior performance of DAST-GCN against baseline models, more predominantly against correlation-based methods which suggest the importance of spatio-temporal modelling of brain activity. This is also supported by the performance of \textit{ST-GCN}, \textit{DAST-GCN\_corr} and the results from \cite{azevedo2020towards} which proves the shortcomings of static pre-defined graph structure based on correlations. The ablation results verify the significance of accounting for the low temporal resolution of the signal before temporal feature extraction. Finally, the comparison of utilizing undirected or single adjacency matrix against $M$ = $K$ dense graphs shows comparable results on sex classifcation and slightly more inferior results on age classification.
Further, to assess the practical applicability and relevance of the proposed model in a clinical setting where the number of labeled samples is often scarce, we conducted the classification experiments for the two tasks on different scales of available data points per class (250, 500, 1000, 2500) and report the 5-fold balanced test accuracy results in Fig. \ref{fig2}.

\begin{table}
\resizebox{\textwidth}{!}{
\begin{tabular}{l|SSS|SSS}
    \toprule
    \multirow{3}{*}{Model} &
      \multicolumn{3}{c}{Sex} &
      \multicolumn{3}{c}{Age } \\
      & { Acc.\%} & { Sens.\%} & {Spec. \% } & { Acc.\%} & {Sens. \%} & {Spec. \%} \\
      \midrule
    SVM-linear & 70.1$\pm$1.9 & 69.9$\pm$2.8 & 74.1$\pm$3.4 & 54.3$\pm$1.3 & 47.7$\pm$1.4 & 55.7$\pm$1.6 \\
    SVM-rbf & 75.1$\pm$1.7 & 77.4$\pm$2.7 & 72.5$\pm$4 & 57.8$\pm$1.2 & 50.1$\pm$1.3 & 66.8$\pm$7 \\
    FCN &79.2$\pm$1.0  & 79.5$\pm$0.7  &80.4$\pm$2.5  &63.1$\pm$1.1  &61.3$\pm$1.5  &61.1$\pm$4.2 \\
    1D-CNN\cite{el2019simple} & 81.7$\pm$1.4 & 82.7$\pm$2.1 & 81.4$\pm$0.8 & 65.9$\pm$0.9& 64.2$\pm$1.1 & 66.4$\pm$2 \\
    BrainNetCNN\cite{kawahara2017brainnetcnn} & 77.8$\pm$1.8  & 78.1$\pm$2.3  & 77.9$\pm$2.9  & 60.9$\pm$1.2  & 60.3$\pm$1.8  &62.5$\pm$1.3  \\
    ST-GCN\cite{gadgil2020spatio} & 78.4$\pm$2.3  & 77.2$\pm$1.9  & 79.8$\pm$1.8  & 60.5$\pm$2.4  & 61.4$\pm$1.9  &60.1$\pm$2.2  \\

    \hline
    \textit{DAST-GCN\_corr} &  78.7$\pm$2.1& 77.5$\pm$1.8  & 79.3$\pm$2.5 & 61.6$\pm$2.9  & 61.1$\pm$3.4  & 62.7$\pm$5.6 \\
    \textit{DAST-GCN\_tlc} & 82.6$\pm$1.3  & 82.3$\pm$1.2   & 82.1$\pm$1.3  &61.4$\pm$2.6  &60.6$\pm$1.7  &61.1$\pm$1.3 \\
    \textit{DAST-GCN\_undir} & 82.9$\pm$1.4  & 83.1$\pm$1.7  &82.9$\pm$1.4  & 64.7$\pm$4.3  &61.3$\pm$2.4  & 63.1$\pm$2.1\\
    \textit{DAST-GCN\_M=1} & 84.8$\pm$0.7 & 84.9$\pm$0.8  & 84.8$\pm$0.7   &64.3$\pm$3.6  & 65.4$\pm$2.7  & 64.3$\pm$3.6 \\

    \hline
    DAST-GCN & 85.3$\pm$0.6 & 85.4$\pm$0.7 & 85.3$\pm$0.7 & 68.6$\pm$1.6 & 67.9$\pm$1.5 & 68.4$\pm$1.6 \\
    \bottomrule
  \end{tabular}
  }
\\
\caption{5-fold test results (mean $\pm$ standard deviation) of DASTG-GCN against baselines and ablated versions of the model.}
\end{table}

\section{Generalizability}
A main challenge of applying deep learning models on neuroimaging datasets is the limited availability of labelled samples and the heterogeneity between fMRI datasets in terms of scanning parameters and participant demographics which restricts the application of pre-trained models across datasets. This explains the wide adoption of linear and shallow non-linear models on engineered features in neuroimaging. However, we hypothesize that a robust graph structure inferred for one task should be the invariant to different scanning parameters and demographics and thus DAST-GCN could be utilized to boost the performance on independent datasets with a limited number of labelled samples.
We validated our hypothesis by transferring the graph structure trained on the UKBiobank dataset for sex classification to 120 healthy participants of the REST-meta-MDD dataset \cite{yan2019reduced} (60 males / 60 females, Age = 40.3 $\pm$15.6 years , TR = 1.3 s, T=220 timepoints) for the same task. We chose the \textit{DA-STGCN\_M=1} version of the model due to the different temporal resolution between datasets and lower number of parameters. We compared the performance of the pre-trained model (DA-STGCN\_pretrained) against a model trained from scratch (DAST-GCN\_no-pretrain), SVM-linear and SVM-rbf baselines. 
Test results of the 5-fold cross validation are reported in Table 2. Accuracy of the pretrained model was more than 7 percent point higher than SVM or without pretraining, with less than half the standard deviation. Our results support that the inferred graph structure is robust across datasets with different scanning parameters and demographics, and highlight the potential advantage of applying DAST-GCN in heterogeneous environments where datasets are collected from different sites.

\begin{table}
\centering
\caption{5-fold test results (mean $\pm$ standard deviation) for sex classification on Rest-meta-MDD dataset.}\label{tab2}
\begin{tabular}{|c|c|c|c|}
\hline
Model  & Acc.\% & Sens.\% & Spec.\% \\
\hline
 Svm-rbf & 61.5$\pm$6.1 & 65.9$\pm$8.6 & 56.6$\pm$20.4  \\
 Svm-linear  & 67.6$\pm$10.2 & 70.7$\pm$13.2 & 66.3$\pm$12.4  \\
 DAST-GCN\_nopretrain & 66.6$\pm$12.1& 69.1$\pm$12.6  & 62.4$\pm$11.6 \\
 DAST-GCN\_pretrained & \textbf{75.0$\pm$4.56} & \textbf{79.6$\pm$8.6} & \textbf{70.0$\pm$10.1}  \\
\hline
\end{tabular}
\end{table}

\section{Limitations}
A good fMRI learning model should not only provide good test metrics but also good explainability in order to advance clinical research. DAST\_GCN infers the underlying graph structure while learning, hence offers a direct visualization of potential connectivity bio-markers for the task in hand. However, given training stochasticity and the dynamic and dense representation of the graph, interpetabillity of the visualizations is challenging. A current possible solution for  more robust visualizations in our framework would be to restrict the graph to be static and sparse. Though such setup hinders the model performance, its could provide clinical insights specifically when used to study psychiatric disorders. Nevertheless more research on explainability techniques for dynamic graphs is required. 

\section{Discussion}

In this work we proposed a novel spatio-temporal graph convolution model for phenotype prediction from fMRI scans. Our experimental results highlight the advantage of learning a dynamic graph structure versus pre-defined correlation based values. Further, we show that the task-inferred graph structure is robust against the heterogeneity of fMRI datasets and can be effectively transferred across different population demographics and scanning parameters. Finally, our results showcase the quantitative and qualitative advantage of applying deep learning models for phenotype prediction in neuroimaging at different scales and advocate for the development of models that incorporate proper inductive bias and operate on minimally pre-processed derivatives of the raw data.

\section*{Acknowledgement}
This work was supported by the Netherlands Organization for Scientific Research (NWO; 628.011.023), Philips Research, AAA Data Science Program, and ZonMW (Vidi; 016.156.318).

%
%
%
\bibliographystyle{splncs04}
\bibliography{main.bib}
\end{document}